\documentclass[journal]{new-aiaa}
\usepackage[utf8]{inputenc}
\usepackage{graphicx,amsmath,booktabs,siunitx}
\usepackage{tikz}
\usetikzlibrary{positioning,fit,calc,backgrounds,arrows.meta}
\usepackage{longtable,tabularx}
\title{An Interpretable Fuzzy Tree for Turbofan Remaining Useful Life with an Auditable Per-Component Diagnostic Layer}

\author{Hugo Henry\footnote{PhD Candidate, Dept.\ of Aerospace Engineering; henryho@mail.uc.edu.}, Bharadwaj Dogga\footnote{PhD Candidate, Dept.\ of Aerospace Engineering; doggabj@ucmail.uc.edu}, Scott Philips\footnote{PhD Candidate, Dept.\ of Aerospace Engineering; philis8@mail.uc.edu}, and Dr. Kelly Cohen\footnote{Professor, Dept.\ of Aerospace Engineering; cohenky@ucmail.uc.edu}}
\affil{University of Cincinnati, Cincinnati, OH 45221, USA}

\begin{document}
\maketitle

\begin{abstract}
Data-driven remaining useful life (RUL) estimators for turbofan engines are accurate but opaque, and their internal features typically encode remaining life rather than any specific component's state. This paper presents an interpretable genetic fuzzy tree that predicts RUL from cruise-condition sensor means on N-CMAPSS while exposing a per-component diagnostic layer inspectable rule by rule. The tree follows the engine hierarchy: sensors to component health, then to bounded component damage, spool damage, and RUL, so each level answers a different maintenance question. Training is staged so the prognostic objective cannot reshape the diagnostic layers: health leaves are fitted to physical targets and frozen, a deterministic fuzzy damage map with fleet-referenced thresholds is imposed on each component, and only the aggregation is optimized for RUL. It consumes 13 cruise-mean scalars per cycle rather than a trajectory, which removes any sequence model. On the native split it reaches 8.44 cycles RMSE on the multi-fault subset and 11.51 across all modes with 333 free parameters, 411 total rules, 32 to 12,000 times fewer than compiled baselines. An exhaustive sensor search bounds which components are diagnosable under the tested cruise-mean design, and the rule base is structured to admit formal verification, which this study does not discharge.
\end{abstract}

\section*{Nomenclature}
{\renewcommand\arraystretch{1.0}
\noindent\begin{longtable*}{@{}l @{\quad=\quad} l@{}}
$x_n$ & cruise-mean sensor vector of cycle $n$, $\in\mathbb{R}^{13}$ \\
$T_n$ & set of cruise samples in cycle $n$ \\
$\theta,\hat\theta$ & component health modifier and its leaf estimate, $\theta\le 0$ \\
$\mu$ & membership function \\
$w_r,\beta_r$ & firing strength and consequent of rule $r$ \\
$d_c,d_s$ & component and spool damage index, $\in[0,1]$ \\
$F_s,F_R$ & spool and root inference maps \\
$p_{10},p_{50},p_{90}$ & fleet quantiles of degrading $\theta$ (damage-map knees) \\
$\phi_L,\phi_C,\phi_P$ & leaf, component, and aggregation parameters \\
$\mathrm{Spec}_x$ & cross-spool specificity \\
$s$ & mean asymmetric prognostic score \\
$\rho$ & per-unit rank correlation \\
$\mathrm{RUL}_{\max}$ & RUL cap (median RUL at degradation onset), cycles \\
\end{longtable*}}
\addtocounter{table}{-1}

\section{Introduction}
\lettrine{A}{n} engine trend alert raised at the gate poses three questions: how much useful life remains, which module is degrading, and whether the answer can support an airworthiness decision. Research has concentrated on the first. Deep temporal networks achieve strong accuracy on N-CMAPSS, the run-to-failure benchmark derived from the Commercial Modular Aero-Propulsion System Simulation, but a scalar remaining useful life (RUL) estimate names no component, and probing the latent features of such models commonly reveals correlation with remaining life rather than with any particular component's state.

The other two carry the operational value. Engines absorb roughly half of airline maintenance expenditure~\cite{mtu_iata}, constrained shop capacity was estimated to add \$5.7 billion to airline engine cost in 2025~\cite{iata2026}, and roughly one signaled fault in six closes as no fault found~\cite{ritter2018}. Removal planning needs a remaining-life estimate, where troubleshooting and workscope definition need a per-component state, both inside the interval between a detectable potential failure and the functional failure~\cite{nowlan1978}.

Engine health monitoring has traditionally been posed as gas path analysis, inferring module efficiency and flow-capacity deviations from steady-state measurements~\cite{urban1973}, a technology surveys describe as more developed in the laboratory than in service~\cite{litt2004}. Fuzzy logic entered early for module-level fault isolation~\cite{ganguli2003,gayme2003} and in operational monitoring scoring each fault on a bounded scale from per-flight cruise snapshots on an airline CFM56 fleet~\cite{demirci2008}, where expert-authored rules detect faults without predicting remaining life. The inverse problem is ill-posed, an engine offering fewer independent measurements than its ten module health parameters~\cite{simon2010}. This work takes those ten as targets and per-cycle cruise means as inputs, learns its rules, and measures which parameters cruise resolves.

The nodes used here are zero-order Takagi and Sugeno systems~\cite{takagi1985}, built on Zadeh's graded membership and linguistic variables~\cite{zadeh1965,zadeh1975,mamdani1975} and tuned by a genetic algorithm~\cite{cordon2001,herrera2008}; cascading them gives the genetic fuzzy tree (GFT)~\cite{ernest2015,ernest2016}. The cascade is what makes the problem tractable: one fuzzy system over all 13 sensors with three membership functions each enumerates $3^{13}=\num{1594323}$ rules, where the tree spans the same sensors with \num{411}.

Structure buys only one half of interpretability. Gacto et al.~\cite{gacto2011} separate complexity, few rules over few variables, from semantics, those variables standing for something. A tree earns the first by construction; the second survives only if protected, since a search optimized end to end against an RUL loss may repurpose the health nodes. An intermediate node is meaningful only with a declared meaning fixed before training, a target of its own, protection from downstream optimization, and demonstrated observability.

The contributions are a four-level GFT with an embedded per-component diagnostic layer and a terminal boundary imposed by construction; a staged scheme that prevents prognostic optimization from altering the diagnostic representation, with its cost quantified; a deterministic damage map whose thresholds do not move with the partition; and an exhaustive sensor study of per-component observability at cruise, including the components for which no assignment succeeds.

\section{Methodology}

\subsection{Dataset}
The study uses the N-CMAPSS run-to-failure dataset~\cite{ariaschao2021}, which simulates degradation under recorded flight conditions, restricted here to cruise. The pooled working set holds 99 engine units and 7473 cycle-level records across nine files: isolated faults of the fan, low- and high-pressure compressor (LPC, HPC) and high- and low-pressure turbine (HPT, LPT), a combined LPC and HPC mode, and the multi-fault DS08 files in which all five degrade. Inputs are the 13 physical sensor channels. Cohen et al.~\cite{cohen2023}, used for comparison in Sec.~III.B, report 90 units over eight subsets, so the populations are similar but not identical.

For each unit and flight cycle $n$ the model consumes one scalar per sensor, the cruise-regime mean
\begin{equation}\label{eq:mean}
x_{n,j}=\frac{1}{|T_n|}\sum_{t\in T_n}s_j(t),\qquad j=1,\dots,13,
\end{equation}
where $T_n$ indexes the cruise samples of cycle $n$. Averaging over $|T_n|\sim10^3$ samples attenuates independent measurement noise and removes the window length, stride and sequence capacity that trajectory methods must select, at the cost of discarding transients. The $1/|T_n|$ variance reduction of independent sampling is an upper bound rather than the realized one, since samples within a cruise segment are autocorrelated; the achieved reduction was not measured. The $x_n$ are residualized against a regression on the operating conditions fitted to each unit's early-life cycles, then smoothed by an exponentially weighted mean of span 7, equivalent to $\alpha=0.25$. This mirrors operational monitoring, where a stabilized cruise snapshot per flight is corrected to reference conditions and smoothed with constants between 0.1 and 0.3~\cite{demirci2008}. Targets are the ten health modifiers $\theta$ supplied by the simulation, flat while a component is healthy and declining after onset. Engine age is not an input, for the reason in Sec.~\ref{sec:limits}. Every fitted preprocessing or calibration step is local to one unit or to one training split, with one disclosed exception; Appendix~\ref{app:prov} gives the data flow.

\subsection{Dataset Assumptions and Their Consequences}\label{sec:assump}
Three properties of the benchmark bound what any model trained on it may claim, and together form the operational domain within which the outputs mean what they are said to mean. Every unit is flown to catastrophic failure, so the fleet holds no censored histories and no survivors, where operational fleets consist largely of such records~\cite{demirci2008}. Every unit carries at least one degrading component, so a healthy reading means only that onset has not been passed. Degradation is monotone and terminal. The terminal boundary of Eq.~\eqref{eq:anchor} transcribes the third property and needs re-derivation wherever a degraded component can be repaired or stabilize, and the damage thresholds are quantiles over degrading trajectories only, well defined because degradation always terminates in failure. A further choice is ours: RUL is capped at $\mathrm{RUL}_{\max}$, the median RUL at onset, computed from training data only. Capping lowers reported RMSE, so comparisons in Sec.~III.B against uncapped studies favor this work due to the introduction of this dataset heuristic bias.

\subsection{Architecture}
\begin{figure}[hbt!]\centering
\resizebox{0.80\textwidth}{!}{%
\begin{tikzpicture}[
  font=\small,
  >={Stealth[length=1.8mm]},
  box/.style={draw, fill=white, align=center, inner sep=2pt},
  leaf/.style={box, minimum width=21mm, minimum height=4.6mm, font=\footnotesize},
  comp/.style={box, minimum width=14mm, minimum height=8mm},
  prog/.style={comp, dashed, fill=black!14},
  spool/.style={box, minimum width=15mm, minimum height=10mm},
  root/.style={box, minimum width=14mm, minimum height=11mm, line width=0.9pt},
  hdr/.style={font=\footnotesize, align=center},
  ttl/.style={font=\footnotesize\bfseries, align=center},
  note/.style={font=\footnotesize, align=center},
  arr/.style={->, line width=0.45pt},
]
\def\xin{-0.1}\def\xleaf{2.75}\def\xcomp{5.75}\def\xspool{8.95}\def\xroot{11.7}
\def\yHPC{0}\def\yHPT{-1.15}\def\yFAN{-2.55}\def\yLPC{-3.70}\def\yLPT{-4.85}
\def\yh{1.05}\def\yt{1.72}\def\ybot{-5.35}

\node[box, minimum width=17mm, minimum height=56mm] (in) at (\xin,-2.42)
  {Cruise-mean\\sensors\\[3pt]$x_n\in\mathbb{R}^{13}$\\[3pt]Eq.~\eqref{eq:mean}};
\node[note, above=3mm of in] (raw) {1\,Hz flight data};
\draw[arr] (raw) -- (in);

\foreach \c/\y in {HPC/\yHPC, HPT/\yHPT, fan/\yFAN, LPC/\yLPC, LPT/\yLPT}{
  \node[leaf] (\c e) at (\xleaf,\y+0.25) {$\hat\theta_{\mathrm{\c,eff}}$};
  \node[leaf] (\c f) at (\xleaf,\y-0.25) {$\hat\theta_{\mathrm{\c,flow}}$};
  \draw[arr] (in.east |- \c e) -- (\c e.west);
  \draw[arr] (in.east |- \c f) -- (\c f.west);
}

\node[comp] (HPC) at (\xcomp,\yHPC) {HPC\\$d_{\mathrm{HPC}}$};
\node[comp] (HPT) at (\xcomp,\yHPT) {HPT\\$d_{\mathrm{HPT}}$};
\node[prog] (fan) at (\xcomp,\yFAN) {fan\\$d_{\mathrm{fan}}$};
\node[prog] (LPC) at (\xcomp,\yLPC) {LPC\\$d_{\mathrm{LPC}}$};
\node[comp] (LPT) at (\xcomp,\yLPT) {LPT\\$d_{\mathrm{LPT}}$};
\foreach \c in {HPC,HPT,fan,LPC,LPT}{
  \draw[arr] (\c e.east) -- ($(\c.west)+(0,0.13)$);
  \draw[arr] (\c f.east) -- ($(\c.west)+(0,-0.13)$);
}

\node[spool] (HP) at (\xspool,{(\yHPC+\yHPT)/2}) {HP spool\\$d_{\mathrm{HP}}$};
\node[spool] (LP) at (\xspool,\yLPC) {LP spool\\$d_{\mathrm{LP}}$};
\draw[arr] (HPC.east) -- ($(HP.west)+(0,0.18)$);
\draw[arr] (HPT.east) -- ($(HP.west)+(0,-0.18)$);
\draw[arr] (fan.east) -- ($(LP.west)+(0,0.25)$);
\draw[arr] (LPC.east) -- (LP.west);
\draw[arr] (LPT.east) -- ($(LP.west)+(0,-0.25)$);

\node[root] (R) at (\xroot,{((\yHPC+\yHPT)/2+\yLPC)/2}) {$\widehat{\mathrm{RUL}}$};
\draw[arr] (HP.east) -- ($(R.west)+(0,0.2)$);
\draw[arr] (LP.east) -- ($(R.west)+(0,-0.2)$);

\node[note, below=1.5mm of LP] (nS) {$d_s{=}1$ if any $d_c{=}1$};
\node[note, below=1.5mm of R]  (nR) {$\widehat{\mathrm{RUL}}{=}0$ if any $d_s{=}1$};

\node[hdr] (hL) at (\xleaf,\yh)  {Health leaves\\3 sensors each};
\node[hdr] (hC) at (\xcomp,\yh)  {Damage map\\Eq.~\eqref{eq:dc}};
\node[hdr] (hS) at (\xspool,\yh) {Spools\\Eq.~\eqref{eq:anchor}};
\node[hdr] (hR) at (\xroot,\yh)  {Root\\Eq.~\eqref{eq:anchor}};
\node[ttl] (tF) at ({(\xleaf+\xcomp)/2},\yt)  {Frozen with respect to RUL};
\node[ttl] (tL) at ({(\xspool+\xroot)/2},\yt) {Optimized on RUL};
\coordinate (bF) at (\xleaf,\ybot);
\coordinate (bL) at (\xspool,\ybot);

\begin{scope}[on background layer]
  \node[fit=(tF)(hL)(hC)(HPCe)(LPTf)(LPT)(bF), fill=black!7, draw=black!35,
        inner xsep=2.5mm, inner ysep=1.5mm] (frozen) {};
  \node[fit=(tL)(hS)(hR)(HP)(LP)(R)(nS)(nR)(bL), draw=black!45, dashed,
        inner xsep=2.5mm, inner ysep=1.5mm] (learned) {};
\end{scope}

\node[note, above=4mm of frozen.north]  (th) {$\theta$, target of the leaves};
\draw[arr, line width=0.8pt] (th.south) -- (frozen.north);
\node[note, above=4mm of learned.north] (ru) {RUL, target of spools and root};
\draw[arr, line width=0.8pt] (ru.south) -- (learned.north);

\node[prog, minimum width=6mm, minimum height=3.2mm, anchor=west] (lg)
  at ($(frozen.south west)+(0,-0.42)$) {};
\node[note, right=1.5mm of lg, align=left]
  {prognostic input only, not independently diagnosable at cruise (Sec.~III.D)};
\end{tikzpicture}}
\caption{Four-level tree. Frozen health leaves fitted to $\theta$ feed the deterministic damage map of Eq.~\eqref{eq:dc}; only the spool and root nodes are optimized against RUL, under the boundary conditions of Eq.~\eqref{eq:anchor}. Dashed components are prognostic inputs only.}
\label{fig:arch}\end{figure}
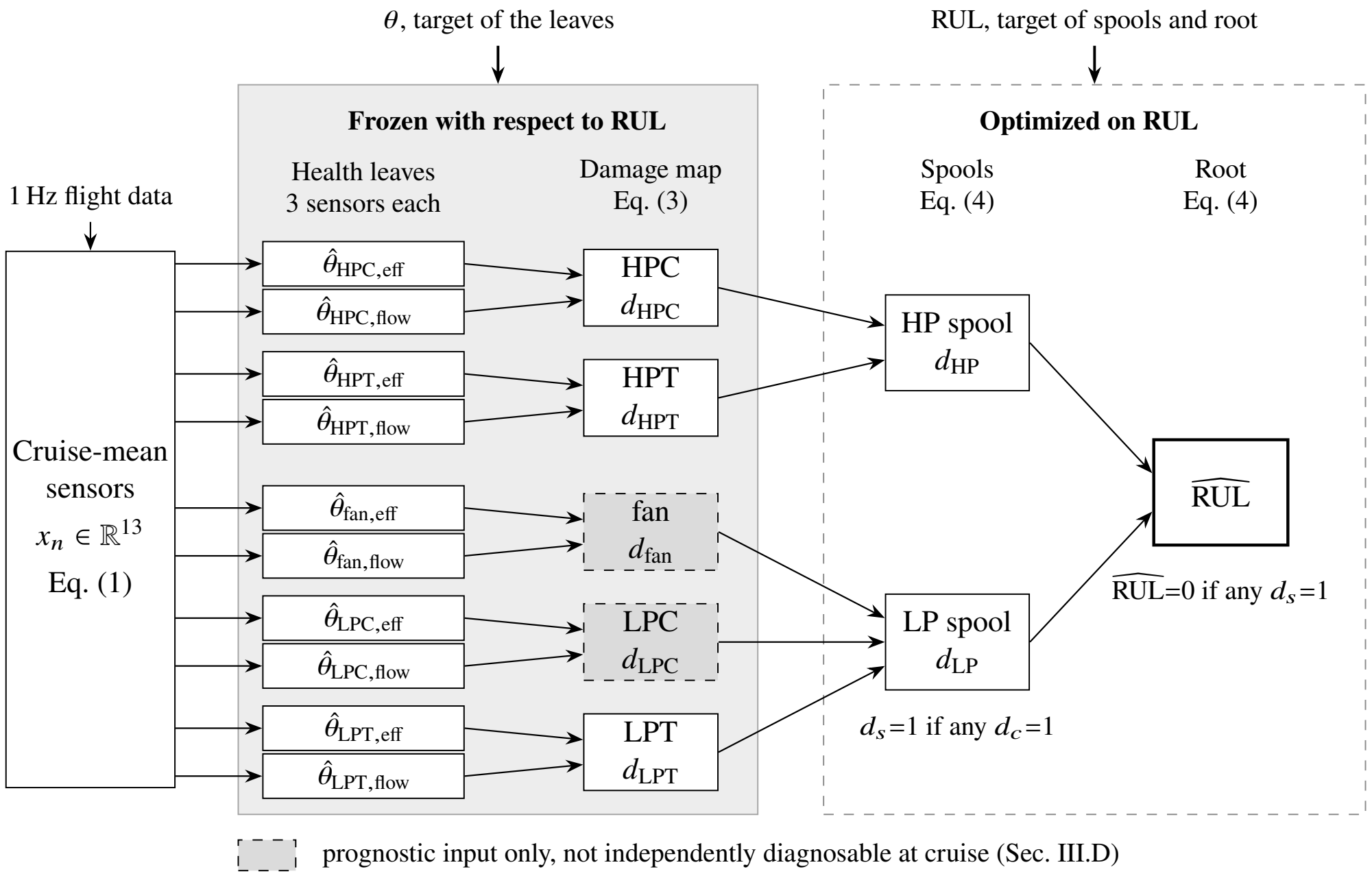

The model is a four-level fuzzy tree evaluated from the sensors upward (Fig.~\ref{fig:arch}), every node a zero-order Sugeno fuzzy inference system (FIS). For inputs $u=(u_1,\dots,u_D)$, each input $j$ carries $k_j$ triangular membership functions at ordered centres forming a partition of unity, $\sum_{i}\mu^{(j)}_i(u_j)=1$. With the product t-norm, the firing strength of rule $r=(i_1,\dots,i_D)$ and the node output are
\begin{equation}\label{eq:fis}
w_r(u)=\prod_{j=1}^{D}\mu^{(j)}_{i_j}(u_j),
\qquad
y(u)=\frac{\sum_r w_r(u)\,\beta_r}{\sum_r w_r(u)}=\sum_r w_r(u)\,\beta_r ,
\end{equation}
the second equality holding because the partition of unity gives $\sum_r w_r\equiv1$. Each output is a convex combination of its consequents, so $y(u)\in[\min_r\beta_r,\max_r\beta_r]$ and every internal signal is bounded by construction rather than by training.

\paragraph{Health leaves.} Leaf $m$ maps a three-sensor subset $S_m$, selected in Sec.~\ref{sec:abl}, to $\hat\theta_m=\sum_r w_r(x_{S_m})\beta^{(m)}_r$ using three membership functions per input at sensor quantiles, giving $3^3=27$ rules, three having shown the best held-out fidelity in a capacity sweep over two to five. The consequents are fitted to $\theta_m$ and held fixed.

\paragraph{Component nodes.} Let $M_c$ be the modifier set of component $c$ and let $p_{10}^{(m)},p_{50}^{(m)},p_{90}^{(m)}$ be quantiles of $\theta_m$ over the degrading trajectories of the training units, the reference population to which every damage value is referred. Each estimate passes through a fixed three-region map and the results combine disjunctively,
\begin{equation}\label{eq:dc}
g_m(\hat\theta)=\mu_{\mathrm{F}}(\hat\theta)+\tfrac12\,\mu_{\mathrm{D}}(\hat\theta),
\qquad
d_c=\max_{m\in M_c} g_m(\hat\theta_m)\in[0,1],
\end{equation}
where $\mu_{\mathrm{F}},\mu_{\mathrm{D}},\mu_{\mathrm{H}}$ are a saturating left trapezoid at $p_{10}^{(m)}$, a triangle at $p_{50}^{(m)}$ and a saturating right trapezoid at $p_{90}^{(m)}$, so $g_m=1$ for $\hat\theta\le p_{10}^{(m)}$ and $g_m=0$ for $\hat\theta\ge p_{90}^{(m)}$. Quantiles of that population rather than of the sample at hand make $d_c$ comparable across units of this fleet. The index is a position within that reference distribution, not a calibrated fraction of physical damage; whether the thresholds carry to another engine or dataset is untested. As rules, each component contributes three with fixed consequents: damage is 1 if either modifier is failed, $\tfrac12$ if either is degrading, and 0 otherwise.

\paragraph{Spool and root.} Damages are grouped by shaft, $C_{\mathrm{HP}}=\{\mathrm{HPC},\mathrm{HPT}\}$ and $C_{\mathrm{LP}}=\{\mathrm{fan},\mathrm{LPC},\mathrm{LPT}\}$, each giving $d_s=F_s(\{d_c\}_{c\in C_s})\in[0,1]$, and the root maps the pair to life, $\widehat{\mathrm{RUL}}=F_R(d_{\mathrm{LP}},d_{\mathrm{HP}})$. Consequent grids are decoded from non-negative increments by cumulative summation, making $F_s$ nondecreasing and $F_R$ nonincreasing in every argument. A disjunctive anchor fixes the boundary conditions
\begin{equation}\label{eq:anchor}
\exists\,c\in C_s:\ d_c=1 \;\Longrightarrow\; d_s=1,
\qquad
\max(d_{\mathrm{LP}},d_{\mathrm{HP}})=1 \;\Longrightarrow\; \widehat{\mathrm{RUL}}=0,
\end{equation}
together with $F_s(0,\dots,0)=0$ and $F_R(0,0)=\mathrm{RUL}_{\max}$. Both are conditions on the \emph{estimated} damage: saturation of some $d_c$ forces $d_s=1$ and saturation of some $d_s$ forces $\widehat{\mathrm{RUL}}=0$. Neither states that a failure the leaves cannot see will drive either to saturation, and Sec.~III.C is the case where it does not. Membership positions in these two nodes are learnable under an order-preserving encoding with pinned endpoints, so the aggregation adapts to the curvature of the damage-to-RUL relationship while the rule base stays readable.

Enumerating every antecedent combination gives \num{411} rules, of which 99 carry a consequent fixed by construction: 45 realize Eq.~\eqref{eq:dc} and 54 are pinned by Eq.~\eqref{eq:anchor}, which determines every cell with a saturated input together with the all-healthy corner. Writing each such set as one rule with a disjunctive antecedent groups the listing without altering the inference and leaves 333 rules, 312 fitted individually and 21 fixed. The 333 free parameters are 270 leaf consequents fitted to $\theta$, 42 aggregation consequents and 21 learnable membership positions; the last two groups, 63 in all, are the only ones optimized against RUL. A further 30 quantile thresholds are estimated from training data rather than optimized, and leaf membership centres are fixed at sensor quantiles. The count is of inference parameters only; Appendix~\ref{app:prov} states what it excludes.

Three rules from the exported base illustrate the form of each level: \textsf{leaf}, IF T30 is High AND P15 is Low AND P40 is Medium THEN $\hat\theta_{\mathrm{HPC,eff}}=-0.00186$; \textsf{component}, IF $\hat\theta_{\mathrm{HPC,eff}}$ is Failed OR $\hat\theta_{\mathrm{HPC,flow}}$ is Failed THEN $d_{\mathrm{HPC}}=1$; \textsf{root}, IF $d_{\mathrm{LP}}$ is Failed THEN $\mathrm{RUL}=0$.

\subsection{Training and Evaluation}
Write $\phi_L$ for the leaf consequents, $\phi_C$ for the component thresholds and $\phi_P$ for the spool and root parameters. Training is staged,
\begin{equation}\label{eq:stages}
\phi_L^{\star}=\arg\min_{\phi_L}\mathcal{L}_{\theta}\!\left(\hat\theta_{\phi_L}(X),\theta\right),
\quad
\phi_P^{\star}=\arg\min_{\phi_P}\mathcal{L}_{\mathrm{RUL}}\!\left(F_{\phi_P}\!\left(d_{\phi_C}(\hat\theta_{\phi_L^{\star}}(X))\right),\mathrm{RUL}\right),
\quad
\frac{\partial\mathcal{L}_{\mathrm{RUL}}}{\partial(\phi_L,\phi_C)}\equiv 0,
\end{equation}
with $\phi_C$ fixed by fleet quantiles and never optimized. The final identity is the protection: the prognostic objective has no authority over the representation producing $\hat\theta$ and $d_c$. This is non-interference, not physical correctness, evidenced separately in Sec.~III.D. A genetic search is used because the rule structure is not differentiable and the constraints act on decoded parameters. An ablation replaces the deterministic map by an RUL-fitted layer to isolate its cost. Training takes 28 minutes on an 11th-generation Intel Core i7-11700F at 2.50~GHz; inference is 18 node evaluations per cycle, stateless and with no sequence buffer.

Three protocols are reported: P1, by-unit cross-validation, used for selection; P2, leave-one-dataset-out (LOSO), withholding a fault mode; and P3, the native split, evaluated once on the frozen configuration. Accuracy is measured by RMSE, $R^2$, per-unit rank correlation $\rho$, and an asymmetric score $s$ penalizing late estimates more heavily, also in the blended form $\tfrac12(\mathrm{RMSE}+s)$ of Ref.~\cite{cohen2023}.

\subsection{Sensor Assignment and Diagnosability}\label{sec:abl}
Assignments are chosen by exhaustive search. For each of the $\binom{13}{3}=286$ sensor triples and each health target, a standalone leaf is fitted and scored on held-out units for observability, the rank correlation between $\hat\theta$ and $\theta$, and for cross-spool specificity,
\[
\mathrm{Spec}_x=1-\frac{\mathbb{E}\!\left[\,|\hat\theta|\;\middle|\;\text{target healthy, opposite spool degrading}\,\right]}
{\mathbb{E}\!\left[\,|\hat\theta|\;\middle|\;\text{target degrading}\,\right]}.
\]
The opposite-spool control separates a component-specific reading from one tracking shaft- or engine-level degradation, and depends on the single-fault subsets where one spool degrades and the other does not. $\mathrm{Spec}_x$ is a ratio of means whose denominator runs over the target's degrading cycles, which are plentiful, and its stability is reported as the standard deviation across splits rather than assumed. Acceptance uses two floors fixed before the confirmation stage, $\rho\ge0.50$ and $\mathrm{Spec}_x\ge0.40$ on held-out units, and accepted modifiers are confirmed across five further splits; Appendix~\ref{app:prov} gives the selection rule, the margin convention of Table~\ref{tab:diag} and the optimism both carry.

\section{Results}
Unless labeled as an ablation, all results use the constrained model with frozen leaves and the deterministic damage layer.

\subsection{RUL Accuracy and Generalization}
\begin{figure}[hbt!]\centering
\includegraphics[width=.64\textwidth]{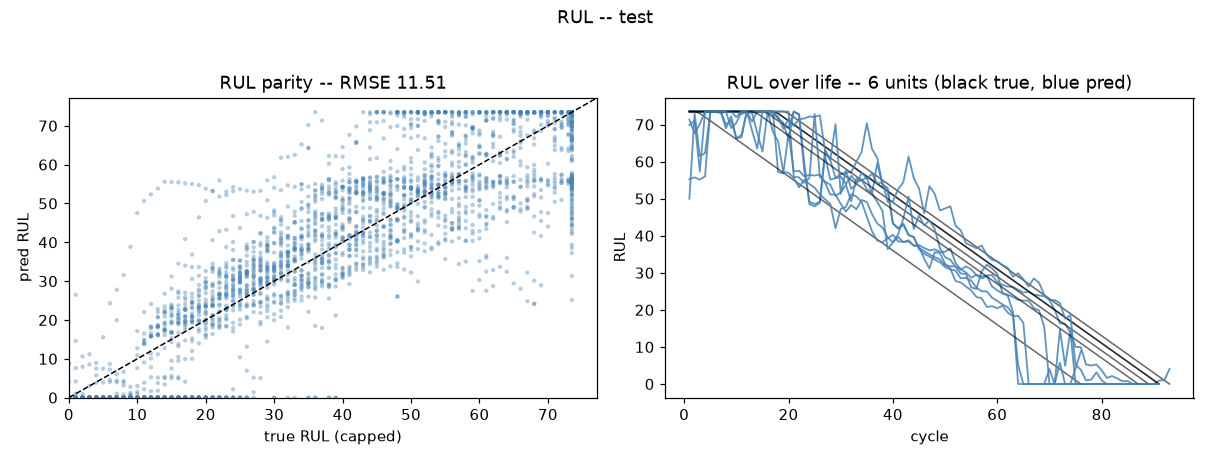}
\caption{Native-test RUL. Left, predicted against true capped RUL. Right, representative per-unit trajectories over life.}\label{fig:rul}\end{figure}

Table~\ref{tab:rul}(a) summarizes accuracy across the three protocols and Fig.~\ref{fig:rul} shows the native-test predictions and representative trajectories. Results sit between 11.5 and 12.2 cycles RMSE whether individual units or entire fault modes are withheld, improving to 8.1 to 8.4 cycles on the multi-fault subset. Standard deviations are across five optimizer seeds on fixed splits, so they describe optimizer variability, not sampling uncertainty over units, of which there are few.

\begin{table}[hbt!]\caption{RUL accuracy. (a) Three protocols; P1 and P2 give mean and standard deviation over five seeds, P3 a single evaluation on the native split. $s$ is the mean asymmetric score, parenthesized values the blended score of Ref.~\cite{cohen2023}. (b) Leave-one-dataset-out, per mode.}\label{tab:rul}\centering\small
(a)\quad\begin{tabular}{llccccc}\toprule
Protocol & Subset & Units & RMSE & $R^2$ & $\rho$ & $s$\\\midrule
P1 (CV)   & DS08 multi-fault & 3/fold  & $8.07\pm1.36$  & 0.764 & 0.939 & 1.03\\
P1 (CV)   & all fault modes  & 11/fold & $11.68\pm1.62$ & 0.732 & 0.920 & 2.41\\
P3 (test) & DS08 multi-fault & 10 & 8.44  & 0.771 & 0.905 & (4.72)\\
P3 (test) & all fault modes  & 39 & 11.51 & 0.735 & 0.922 & (6.75)\\
P2 (LOSO) & mean over modes  & 99 & 12.20 & 0.660 & 0.911 & 2.61\\\bottomrule
\end{tabular}

\vspace{5pt}
(b)\quad\begin{tabular}{lccc}\toprule
Held-out mode & RMSE & $R^2$ & $\rho$\\\midrule
DS01, DS02, DS03 & 10.19--10.80 & 0.75--0.80 & 0.92--0.95\\
DS05, DS06, DS07 & 10.60--12.16 & 0.73--0.78 & 0.93\\
DS08a, DS08c     & 11.53, 13.24 & 0.68, 0.34 & 0.93--0.95\\
\textbf{DS04 (isolated fan)} & \textbf{19.59} & \textbf{0.315} & 0.723\\\midrule
Mean             & 12.20 & 0.660 & 0.911\\\bottomrule
\end{tabular}\end{table}

\subsection{Comparison with Reported Results}
Table~\ref{tab:base} places the model against N-CMAPSS results compiled in Ref.~\cite{cohen2023}. The present model carries 333 free parameters, against \num{10631} for the most compact prior entry, \num{169401} for the deep Gaussian process, and \num{1.03e6} and \num{4.09e6} for the two convolutional models, ratios of $32{\times}$, $509{\times}$, $3.1{\times}10^{3}$ and $1.2{\times}10^{4}$. Counting only the 63 parameters the prognostic objective adjusts, the ratio to the most compact prior entry is $169{\times}$. Part of the reduction comes from Eq.~\eqref{eq:mean}: 13 scalars per cycle carry no convolutional stack, recurrent state or sequence embedding. Accuracy falls within the range of the compiled studies, and this is the only entry exposing a continuous per-component health state. No ranking is intended: the rows differ in subset, split, features and capping, no prior model was rerun under our protocol, and only the all-modes native-split row is approximately matched, that one still differing in features and capping. The rest is literature context.

\begin{table}[hbt!]\caption{Reported N-CMAPSS RUL RMSE and model size, values other than ours as compiled in Ref.~\cite{cohen2023}. Subset, split, features and capping differ across rows; two further entries reporting 6.87 and 5.64 cycles state no parameter count and are omitted.}\label{tab:base}\centering\footnotesize
\begin{tabular}{llrcl}\toprule
Approach & Evaluation & Parameters & RMSE & Diagnostic output\\\midrule
DeVol et al.~\cite{devol2021}                   & native split, context & \num{1030000} & 12.5 & none\\
Sol\'is-Mart\'in et al.~\cite{solismartin2021}  & native split, context & \num{4089465} & 6.24 & none\\
Biggio et al.~\cite{biggio2021}                 & subset, context       & \num{169401}  & 7.31 & none\\
Cohen et al.~\cite{cohen2023} (ANN--Flux+PCA)   & native split, nearest & \num{10631}   & $8.20\pm0.17$ & failure classification\\
\textbf{GFT (this work)}      & native, all modes & 333 & 11.51 & per-component index\\
GFT (this work)               & native, DS08 subset & 333 & 8.44 & per-component index\\\bottomrule
\end{tabular}\end{table}

\subsection{Where Generalization Fails}
Per-mode LOSO results appear in Table~\ref{tab:rul}(b). Errors are uniform at roughly 10 to 13 cycles with one exception, the isolated-fan mode DS04, which reaches 19.59 cycles with $R^2=0.315$ and alone lifts the mean to 12.20. The sensor study of Sec.~III.D predicts this, since the fan is not independently observable from cruise means. The failure is legible: the damage index rises only part way while true RUL declines, so the estimate stays near the cap and errs optimistically, where a model without an explicit damage state returns the same value with no sign the fault went unread. Rank correlation on these units is still 0.72, so the estimate orders the trajectory roughly correctly while being badly scaled; rank agreement and calibrated error are separate claims and a removal decision needs the second.

\subsection{Per-Component Diagnosability}
\begin{figure}[hbt!]\centering
\includegraphics[width=.55\textwidth]{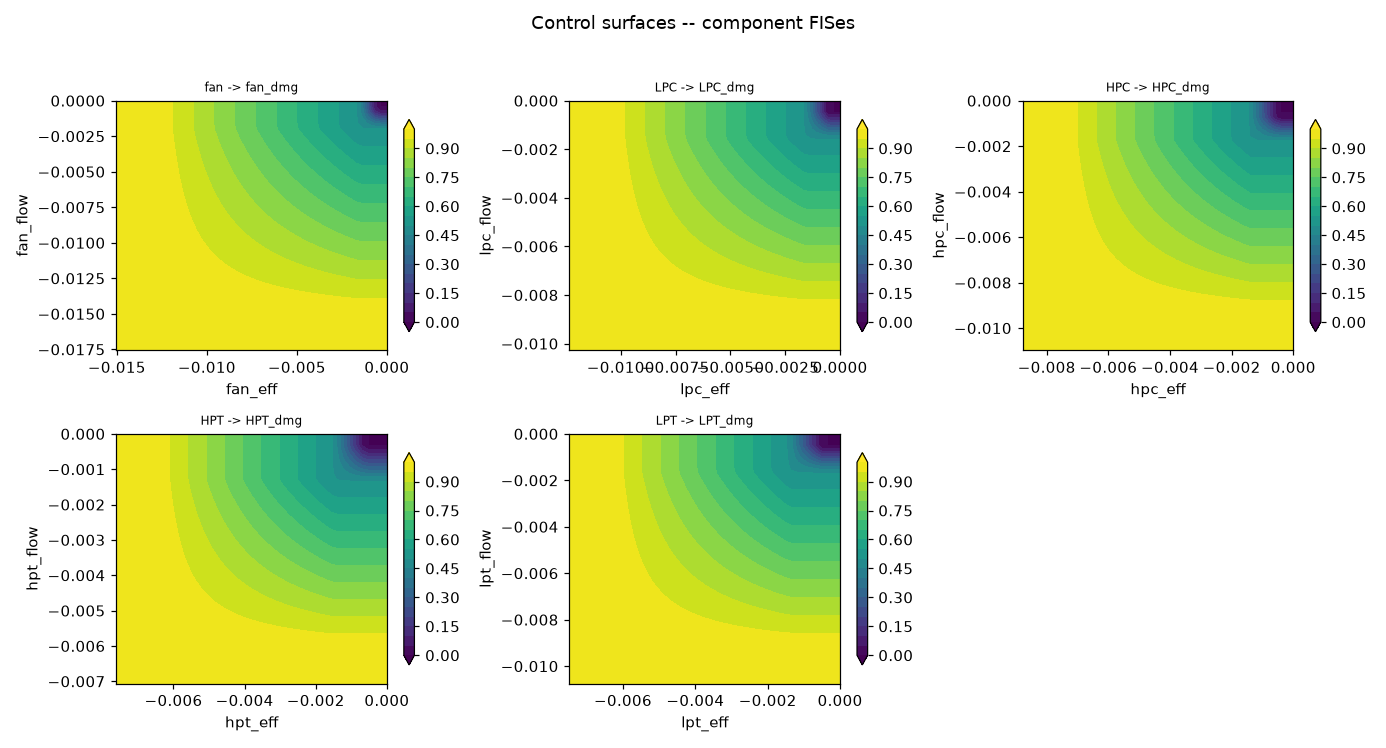}
\caption{Deterministic component damage maps from $\hat\theta$ to $d_c$, saturating beyond the fleet thresholds of Eq.~\eqref{eq:dc}. The knees are quantiles of the degrading training units, so the map does not move with the evaluation split.}\label{fig:surf}\end{figure}

Table~\ref{tab:diag} reports the exhaustive search after confirmation over five splits. Six of the ten modifiers clear both floors of Sec.~\ref{sec:abl}, but not equally. Both HPC modifiers and LPT efficiency clear the specificity floor by more than one standard deviation across splits and are reported as diagnostic; both HPT modifiers and LPT flow clear it by one or less and are reported as marginal, a weaker claim than the floor alone suggests. No triple makes the fan or LPC clear both floors, so they are retained as prognostic inputs only, consistent with the ill-conditioning of steady-state gas path analysis, where module health parameters outnumber the independent measurements~\cite{simon2010,ganguli2003}. Figure~\ref{fig:surf} shows the resulting damage maps.

\begin{table}[hbt!]\caption{Per-component diagnosability from exhaustive three-sensor search, scored on held-out units, mean and standard deviation over five splits. Acceptance floors are $\rho\ge0.50$ and $\mathrm{Spec}_x\ge0.40$; margin is $(\mathrm{Spec}_x-0.40)$ in units of that standard deviation, so values at or below 1 clear the floor by no more than its own spread. Fan and LPC rows span both modifiers' best triples; LPC efficiency was rejected at the single-split screening stage and was not carried into the five-split confirmation.}\label{tab:diag}\centering\footnotesize
\begin{tabular}{llcccl}\toprule
Modifier & Sensors & $\rho$ & Cross-spool spec. & Margin & Verdict\\\midrule
HPC eff.  & T30, P15, P40 & 0.77 & $0.74\pm0.08$ & 4.3 & diagnostic\\
HPC flow  & T30, P15, P21 & 0.71 & $0.66\pm0.11$ & 2.4 & diagnostic\\
LPT eff.  & Nc, T30, T48  & 0.77 & $0.54\pm0.11$ & 1.3 & diagnostic\\
HPT flow  & Wf, Nc, T30   & 0.72 & $0.60\pm0.20$ & 1.0 & marginal\\
HPT eff.  & Nc, Ps30, P50 & 0.74 & $0.50\pm0.13$ & 0.8 & marginal\\
LPT flow  & Nc, T30, T48  & 0.90 & $0.42\pm0.07$ & 0.3 & marginal\\
LPC (both)& none of 286 passes & 0.48--0.70 & 0.11--0.32 & below floor & prognostic only\\
fan (both)& none of 286 passes & 0.44--0.72 & 0.15--0.40 & below floor & prognostic only\\\bottomrule
\end{tabular}\end{table}

\subsection{Cost of the Diagnostic Constraint}
Replacing the deterministic component layer with an RUL-fitted one lowers held-out DS08 RMSE from 8.07 to 6.78 cycles, so the constraint costs about 1.3 cycles against marginal means whose seed standard deviations overlap. The comparison is paired by construction: for a given seed both arms are trained and scored on the identical unit partition and differ in nothing but the component layer, so pairing on held-out unit and averaging within unit across the five seeds removes the split-to-split variance that dominates those marginals. The paired mean difference, with a bootstrap interval clustered on units rather than on the correlated unit-by-seed observations, a paired comparison over the 11 held-out units common to both arms, averaged within unit across 5 seeds, gives a mean difference of 2.14 cycles (95\% unit-clustered bootstrap interval 1.13 to 3.34). The unconstrained variant is more accurate, but its component states are shaped by the RUL loss and no longer support a per-component health claim.

\section{Discussion}
Accuracy is stable across interpolation to unseen units, the native split and transfer to an unseen mode, and degrades where the sensor study predicts; because the architecture carries an explicit damage state, that limit appears in the damage trajectory rather than inside a confident prediction.

A physical mechanism is consistent with the boundary. The simulation regulates corrected fan speed through fuel flow in response to the throttle-resolver angle~\cite{frederick2007}, so at steady cruise the controller holds fan speed and redistributes low-pressure degradation into fuel flow and temperatures that other faults also move~\cite{demirci2008}. Fan speed appears in none of the ten assignments of Table~\ref{tab:diag}. This is an interpretation consistent with the data, not a tested result.

\subsection{Reuse of the Protected Layers}
Each level was kept from being reshaped after training, so each remains readable on its own and suggests a use: trend monitoring at the leaves, module ranking at the component indices, core against low-pressure system at the spools, removal planning at the root. Those are potential uses; no maintenance decision, isolation error rate or time saving is evaluated here. Against the four requirements of Sec.~I, declared meaning holds by construction, each leaf has its own target $\theta$ and is frozen after fitting, and protection is the identity of Eq.~\eqref{eq:stages}. Observability alone is not uniform: it is established for HPC and for LPT efficiency, marginal for HPT and LPT flow, and absent for the fan and LPC, which are reported at spool level. Reuse of a frozen diagnostic layer by a different prognostic head is a hypothesis this construction makes testable, at the cost of the 1.3 cycles of Sec.~III.E. What the construction establishes is fidelity to the simulator's health modifiers and discrimination under the tested sensor design; operational diagnostic validity is a third claim that this dataset cannot settle.

\subsection{Amenability to Formal Verification}\label{sec:verif}
Arnett et al.\ converted a fuzzy controller to piecewise polynomials and discharged a stability-related property with a satisfiability modulo theories solver~\cite{arnett2017}, then verified a genetic fuzzy system for unmanned aerial vehicle navigation within a safety corridor~\cite{arnett2020}. The present architecture satisfies their structural conditions: the rule base is finite and enumerable, 333 rules grouped and \num{411} expanded; inference is deterministic, feedforward and stateless, so a cycle is a pure function of 13 scalars; and membership functions are piecewise linear with constant consequents, so each output in Eq.~\eqref{eq:fis} is piecewise polynomial and the tree is encodable for satisfiability or mixed-integer analysis without approximation.

Four design choices then become candidate proof obligations rather than empirical observations: $0\le d_c\le 1$ and $0\le\widehat{\mathrm{RUL}}\le\mathrm{RUL}_{\max}$, the terminal implication of Eq.~\eqref{eq:anchor}, monotonicity of $d_s$ and $\widehat{\mathrm{RUL}}$, and coverage. Each holds under stated conditions rather than unconditionally: boundedness requires the partition of unity and consequents inside the declared range; monotonicity requires the cumulative-sum decoding; the terminal implication is a designed boundary on the estimated damage and says nothing about a failure the leaves do not register. No certification claim is made and no solver was run, so none of the four is a discharged proof; the contribution is membership in a class where they could be discharged, which an unconstrained network of $10^6$ parameters is not.

That distinction has a destination. Maintenance credit for health monitoring is granted through an aircraft health monitoring analysis within MSG-3, where on-ground algorithms are validated for effectiveness and reliability by the operator's authority and credit is bounded to non-safety tasks~\cite{imrbpb,faa2022,sae}; assurance of the analytics is still being written~\cite{easa,faa_ai,rajamani2025}. The expectations binding a model there, a defined operational domain with detection of inputs outside it, a mapped region of degraded performance, and explanations usable by maintenance staff, are addressed in Secs.~\ref{sec:assump}, III.C and IV.A.

\subsection{Limitations}\label{sec:limits}
Three limitations follow from Eq.~\eqref{eq:mean}. Cruise-only observation never sees the engine under high load, so a modifier latent at a stabilized cruise point but evident at takeoff power lies outside these features, and the boundary of Sec.~III.D is about cruise rather than about the components. Averaging discards within-cycle dispersion, and the smoothing of Sec.~II.A removes further variability between cycles. What remains is a pure function of 13 scalars, which makes the verification argument straightforward but leaves the inference without accumulated evidence or recursive filtering.

Raw engine age was withheld deliberately, not because it is illegitimate. In a fleet where every unit is flown to failure age is a strong chronological prior, so a model given it can score well without reading degradation at all, leaving the reported accuracy unattributable to the health pathway; the relation also depends on the simulation and does not survive censoring or repair. The specificity test of Sec.~\ref{sec:abl} is bounded similarly: it is a cross-spool control, showing that a retained leaf is not reading engine-level or opposite-shaft degradation, not that two components on one shaft can be told apart. HPC and HPT were never tested against one another, so attribution within a spool is unsupported.

One preprocessing step is not causal. The per-unit condition regression is fitted on the first fifth of that unit's cycles, a window whose length depends on the unit's total life and is not knowable early in that life. No other unit enters the fit and no future sensor value enters the regression, so what the step borrows from the future is the window length alone; Appendix~\ref{app:prov} gives the full flow and the deployed alternative, whose effect on the reported numbers was not measured.

The model returns a point estimate with no predictive interval, where Biggio et al.\ provide one~\cite{biggio2021}. Optimistic errors concentrate on the modes not observable at cruise, where the estimate should be treated as low confidence, flagged by the plateauing damage index. Conclusions are bounded by cruise-mean features, three-sensor leaves, a capped target and the prescribed trajectories.

\section{Conclusion}
\looseness=-1
A genetic fuzzy tree separating diagnostic from prognostic optimization produces RUL estimates of 8.44 cycles on the multi-fault subset and 11.51 across all modes on the native split, from 13 scalars per cycle and 333 rules, 63 of whose parameters the prognostic objective adjusts. Freezing the leaves and imposing a deterministic damage interface guarantees that the objective cannot redefine the internal health states, which is non-interference rather than proof that they are physically right, at a cost near 1.3 cycles. An exhaustive sensor search bounds what may be claimed under the tested three-sensor cruise-mean design: HPC and LPT efficiency clear the acceptance floors by more than their spread across splits, HPT and LPT flow clear them marginally, and no triple makes the fan or LPC clear them at all. Of the three questions of Sec.~I it answers the first with a bounded, monotone estimate, answers the second for part of the engine, and does not answer the third, offering a structure that admits proof rather than a proof. Extension beyond cruise, and beyond a simulation in which every unit fails, is the next step.

\appendix
\section{Data Provenance and Split Boundaries}\label{app:prov}
Every fitted preprocessing or calibration step is local to one unit or to one training split. The operating-condition regression is fitted per unit on that unit's own early cycles, so no other unit enters it and nothing crosses the train and test boundary. The exponentially weighted mean is causal, using only past cycles, so no future measurement enters a prediction. The damage knees of Eq.~\eqref{eq:dc} and the cap $\mathrm{RUL}_{\max}$ are estimated on the training units of each split alone and applied unchanged to the held-out units.

The one departure concerns the length of the regression window, which is the first fifth of a unit's cycles. That fraction is of the unit's total life, which is not knowable at the time of an early-life prediction, so the window length is information the model would not have in service. No other unit's data and no future sensor value enter the fit itself, so the departure is not cross-unit leakage and does not transfer label information across the evaluation boundary. A deployed version would fix the window at a set number of early cycles; the reported numbers were not recomputed under that substitution, and doing so is the first item of future work.

Among the triples clearing both floors, the one with the largest $\mathrm{Spec}_x$ is retained; where none clears both, the triple with the largest $\rho$ is retained and the modifier is marked prognostic only. The parameter count of Sec.~II.C is of inference parameters: it excludes the per-unit preprocessing regressions, the quantile thresholds, which are estimated rather than optimized, and the $286\times10$ candidate leaves fitted during selection, so it measures the deployed model rather than the cost of reaching it. The specificity floors of Sec.~\ref{sec:abl} were fixed before the confirmation stage, and the margin column of Table~\ref{tab:diag} exists because a mean that clears a floor by less than its own spread across splits is not a stable acceptance. Each retained triple is the best of 286 for its target, so both its specificity and its margin are optimistic.

Sensor assignments were selected on held-out units and confirmed on five further splits, but selection and reporting draw on the same fleet, so the accepted specificities of Table~\ref{tab:diag} are maxima over a large candidate set. The complete remedy is nested selection, choosing assignments on units never used for reporting; it was not run here. The negative findings for the fan and the low-pressure compressor are unaffected, since a maximum over 286 candidates that still fails both floors is evidence of absence rather than of selection.

All code used throughout this study and the trained tree parameters are available on the public github repository associated~\cite{repo}.
\section*{Funding Sources}
The authors received no external financial support for the research or authorship.

\section*{Acknowledgments}
The authors would like to thank the members of the AI Bio Lab at the University of Cincinnati for their feedback and discussions throughout this work. The authors also acknowledge Dr. Chetan Kulkarni
(KBR Technical Fellow, Diagnostics and Prognostics Group, 
NASA Ames Research Center), Prof. (asst. TUD) Dr. Manuel Arias Chao (ZHAW School of Engineering IDS Institute for Data Science) and Joseph (Yossi) Cohen (Rutgers School of Engineering) for their valuable feedback and contributions. In particular, Dr. Cohen's work provided an important foundation for this research.

AI-based tools were used to assist with grammar and syntax editing, as well as software development under detailed specifications provided by the authors. This included assistance with the implementation and adaptation of the genetic algorithm to the specific chromosome representation used in this work. All generated or modified material was subsequently reviewed, validated, and revised by the authors over the course of the project. The authors retain full responsibility for the methodology, implementation, analysis, and conclusions presented in this paper.

\end{document}